\documentclass[
]{ceurart}

\usepackage[normalem]{ulem}

\begin{document}

\copyrightyear{2020}
\copyrightclause{Copyright for this paper by its authors.
  Use permitted under Creative Commons License Attribution 4.0
  International (CC BY 4.0).}

\conference{Dravidian-CodeMix-HASOC 2020: Sub-track of Hate Speech and Offensive Content Identification in Indo-European Languages at FIRE 2020}

\title{Gauravarora@HASOC-Dravidian-CodeMix-FIRE2020: Pre-training ULMFiT on Synthetically Generated Code-Mixed Data for Hate Speech Detection}

\author[]{Gaurav Arora}[
email=gaurav@haptik.ai,
url=https://goru001.github.io/,
]
\address[]{Jio Haptik Technologies Limited}

\begin{abstract}
This paper describes the system submitted to Dravidian-Codemix-HASOC2020: Hate Speech and Offensive Content Identification in Dravidian languages (Tamil-English and Malayalam-English). The task aims to identify offensive language in code-mixed dataset of comments/posts in Dravidian languages collected from social media. We participated in both Sub-task A, which aims to identify offensive content in mixed-script (mixture of Native and Roman script) and Sub-task B, which aims to identify offensive content in Roman script, for Dravidian languages. In order to address these tasks, we proposed pre-training ULMFiT on synthetically generated code-mixed data, generated by modelling code-mixed data generation as a Markov process using Markov chains. Our model achieved 0.88 weighted F1-score for code-mixed Tamil-English language in Sub-task B and got 2nd rank on the leader-board. Additionally, our model achieved 0.91 weighted F1-score (4th Rank) for mixed-script Malayalam-English in Sub-task A and 0.74 weighted F1-score (5th Rank) for code-mixed Malayalam-English language in Sub-task B.
\end{abstract}


\maketitle

\section{Introduction}

In recent years, as more and more people from all walks of life come online, it has become very important to monitor their behaviour in order to flag behaviour that is hateful, offensive or encourages violence towards a person or group based on something such as race, religion, sex, or sexual orientation, to ensure Internet remains an inclusive place and to promote diversity of content, thoughts and encourage creativity. It is a challenging task to develop systems that can effectively flag hateful and offensive content accurately \citep{kumar-etal-2018-benchmarking}. Additionally, a substantial amount of work has been done for hate speech detection in languages like English, but no work has been done for Indic languages \citep{hasocdravidian-ceur},  \citep{hasocdravidian--acm}.

Dravidian-Codemix-HASOC2020: Hate Speech and Offensive Content Identification in Dravidian languages (Tamil-English and Malayalam-English) proposes to bridge this gap. It's goal is to identify the offensive language in code-mixed dataset of comments/posts in Dravidian languages collected from social media. Each comment/post is annotated with offensive/not-offensive label at the comment/post level. The data set has been collected from YouTube comments and tweets. It has 2 sub-tasks. Sub-task A is a message-level label classification task in which given a YouTube comment in Code-mixed (Mixture of Native and Roman Script) Malayalam, systems have to classify it into offensive or not-offensive. Sub-task B is also a message-level label classification task in which given a tweet or YouTube comment in Tanglish or Manglish (Tamil and Malayalam written using Roman Script), systems have to classify it into offensive or not-offensive \citep{hasocdravidian-ceur},  \citep{hasocdravidian--acm}.

In this competition, we participated in both sub-task A and sub-task B. We pre-trained ULMFiT \citep{DBLP:journals/corr/abs-1801-06146} on synthetically generated code-mixed data and used transfer learning to train the model on downstream task of hate-speech classification. We propose a Markov model to synthetically generate code-mixed data from Wikipedia articles in native form and their translated and transliterated versions. We use this synthetically generated code-mixed data for Malayalam and Tamil to pre-train ULMFiT from scratch. With this approach, we achieve 2nd Rank on the leader-board for code-mixed Tamil-English language in Sub-task B, 4th Rank on the leader-board for code-mixed Malayalam in mixed-script in Sub-task A and 5th Rank on the leader-board for code-mixed Malayalam-English language in Sub-task B. We've open-sourced our code for dataset preparation, language model pre-training and classification model training on GitHub\footnote{https://github.com/goru001/nlp-for-tanglish}\textsuperscript{,}\footnote{https://github.com/goru001/nlp-for-manglish}.

\section{Related Work}

Lot of researchers and practitioners from industry and academia have been attracted towards the problem of automatic identification of hateful and offensive speech. Authors in \citep{Fortuna2018ASO} discuss the complexity of the concept of hate speech and its unquestionable potential for societal impact, particularly in online communities and digital media platforms. There have been previous attempts at developing models for hate speech detection in English, Hindi and German \citep{10.1145/3368567.3368584}, Italian \citep{10.1145/3377323} but not much work has been done for Dravidian code-mix languages. Having said that, attempts are being made to accelerate progress in NLP in Dravidian languages, like in this competition or in \citep{dravidiansentiment-ceur},
\citep{dravidiansentiment-acm}.

There have been attempts at synthetically generating code-mixed data based on linguistic theory \citep{inproceedings}. But in this paper, we use a novel technique to generate synthetic code-mixed data and pre-train ULMFiT model for Dravidian languages.

\section{Methodology}

In this section we discuss data preparation for pre-training ULMFiT language model, get insights into the Dravidian code-mix HASOC classification train, dev and test sets and discuss modeling details.

\subsection{Dataset}

\begin{figure}
  \centering
  \includegraphics[scale=0.7]{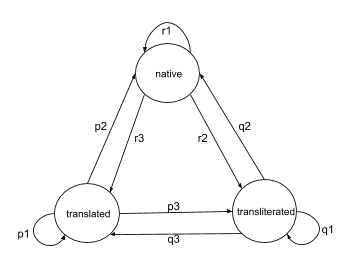}
  \caption{Markov chain modelling code-mixed data generation process}
  \label{markov-fig}
\end{figure}

\textbf{Data preparation for pre-training ULMFiT language model.} Firstly, we obtain parallel sets of Wikipedia articles in native, transliterated and translated form. This gives us one to one mapping between sentences in native, transliterated and translated form. This allows us to model code-mixed data generation process as a Markov Process with 3 states; \texttt{native, translated} and \texttt{transliterated} with \texttt{p1, p2, p3, q1, q2, q3, r1, r2, r3} as state transition probabilities as shown in Figure \ref{markov-fig}. We sample from this Markov process from time \texttt{T=0} to time \texttt{T=N}, where \texttt{N=total no. of sentences in all the Wikipedia articles}, which gives us a output sequence of states $S_{output}$=\{native, transliterated, transliterated, native, translated, .......\}  depending upon the values of state transition probabilities. Using $S_{output}$, we form code-mixed data by picking $i^{th}$ sentence at $T=i$ from $S_{output}$'s $i^{th}$ state. Example of state transition probabilities used for synthetically generating Malayalam code-mixed data are shown in Table \ref{tab:mal-syn-prob}.

\begin{table}[]
\centering
\caption{State transition probabilities of Markov model used for synthetically generating Malayalam code-mixed data   }
\label{tab:mal-syn-prob}
\begin{tabular}{cccccc|cccccc}
\multicolumn{6}{c|}{\textbf{Model 1}} & \multicolumn{6}{c}{\textbf{Model 2}} \\ \hline
p1 & 0 & q1 & 0 & r1 & 0 & p1 & 0 & q1 & 0 & r1 & 0 \\
p2 & 1 & q2 & 0 & r2 & 1 & p2 & 0 & q2 & 1 & r2 & 0 \\
p3 & 0 & q3 & 1 & r3 & 0 & p3 & 1 & q3 & 0 & r3 & 1
\end{tabular}
\end{table}

\textbf{Dravidian code-mix HASOC classification dataset.} Table \ref{tab:data-statistics} shows statistics of the dataset used in this competition for Sub-task A and Sub-task B. Dataset statistics are largely consistent across train, dev and test sets and contain unbalanced classes, varied length sentences. Malayalam and Tamil training datasets for sub-task B contain a few mixed script sentences. 

\begin{table}[]
\centering
\caption{HASOC Dravidian code-mix Dataset statistics}
\label{tab:data-statistics}
\begin{tabular}{@{}cccccccc@{}}
\toprule
\textbf{} & \multicolumn{3}{c}{\textbf{Sub-task A}} & \multicolumn{4}{c}{\textbf{Sub-task B}} \\ \cmidrule(l){2-8} 
\textbf{} & \multicolumn{3}{c}{\textbf{Malayalam mixed-script}} & \multicolumn{2}{c}{\textbf{Malayalam}} & \multicolumn{2}{c}{\textbf{Tamil}} \\ \cmidrule(l){2-8} 
no. of classes & \multicolumn{3}{c}{2} & \multicolumn{2}{c}{2} & \multicolumn{2}{c}{2} \\ \midrule
\textbf{} & \textbf{Train} & \textbf{Dev} & \textbf{Test} & \textbf{Train} & \textbf{Test} & \textbf{Train} & \textbf{Test} \\ \cmidrule(l){2-8} 
no. of examples & 3200 & 400 & 400 & 4000 & 951 & 4000 & 940 \\ \cmidrule(r){1-1}
\begin{tabular}[c]{@{}c@{}}\%age of examples containing\\  only Roman characters\end{tabular} & 66.8\% & 70.2\% & 64.5\% & 99.1\% & 100\% & 99.7\% & 100\% \\ \cmidrule(r){1-1}
min. no. of examples in a class & 567 & 72 & - & 1952 & - & 1980 & - \\ \cmidrule(r){1-1}
max. no. of examples in a class & 2633 & 328 & - & 2047 & - & 2020 & - \\ \cmidrule(r){1-1}
avg. no. of examples in a class & 1600 & 200 & - & 2000 & - & 2000 & - \\ \cmidrule(r){1-1}
min no. of tokens in an example & 1 & 2 & 1 & 1 & 1 & 1 & 1 \\ \cmidrule(r){1-1}
max no. of tokens in an example & 186 & 56 & 186 & 62 & 526 & 66 & 55 \\ \cmidrule(r){1-1}
avg no. of tokens in an example & 9.009 & 9.44 & 9.87 & 9.99 & 10.47 & 18.33 & 17.31 \\ \cmidrule(r){1-1}
median no. of tokens in an example & 8 & 8 & 8 & 8 & 9 & 16 & 15 \\ \bottomrule
\end{tabular}
\end{table}

\subsection{Modeling Details}

\subsubsection{Model Architecture}
Weight-dropped AWD-LSTM variant \citep{DBLP:journals/corr/abs-1708-02182} of the long short-term
memory network \citep{10.1162/neco.1997.9.8.1735} and \citep{10.1162/089976600300015015} was used for the language modeling task. Its embedding size was 400, number of hidden activations per layer was 1152 and the number of layers was 3. Two linear blocks with batch normalization and dropout were added as custom head for the classifier with rectified linear unit activations for the intermediate layer and a softmax activation at the last layer \citep{DBLP:journals/corr/abs-1801-06146}. Table \ref{tab:awd-dropout} shows dropout applied to AWD-LSTM model. We use \textit{Dropout Multiplicity} as a parameter in our experiments to proportionately change the magnitude of all dropouts.

\begin{table}[]
\centering
\caption{Dropout applied to AWD-LSTM model}
\label{tab:awd-dropout}
\begin{tabular}{ccccc}
\hline
\textbf{Layer} & \begin{tabular}[c]{@{}c@{}}Embedding\\  Layer\end{tabular} & \begin{tabular}[c]{@{}c@{}}Input\\ Layer\end{tabular} & \begin{tabular}[c]{@{}c@{}}LSTM's\\ Internal\end{tabular} & \begin{tabular}[c]{@{}c@{}}Between\\ LSTMs\end{tabular} \\ \hline
\textbf{Dropout} & 0.02 & 0.1 & 0.2 & 0.2 \\ \hline
\end{tabular}
\end{table}

\subsubsection{Pipeline}
\textbf{Tokenization.} We create subword vocabulary by training a SentencePiece\footnote{https://github.com/google/sentencepiece} tokenization model on the dataset prepared, using unigram segmentation algorithm \citep{DBLP:journals/corr/abs-1808-06226}. Since we're using subword tokenization, our models can handle different spellings in transliteration or spelling errors in test set. Table \ref{tab:vocab-size} shows vocabulary size of the models trained for each one of the sub-tasks.

\begin{table}[]
\centering
\caption{Vocab size of the model}
\label{tab:vocab-size}
\begin{tabular}{cccc}
\hline
\textbf{Language} & \textbf{Tamil} & \multicolumn{2}{c}{\textbf{Malayalam}} \\ \hline
\textbf{Script} & Latin & Latin & Mixed \\
\textbf{Vocab size} & 8000 & 15000 & 25000 \\ \hline
\end{tabular}
\end{table}

\textbf{ULMFiT Language model pre-training.} Our model is based on the Fastai\footnote{https://github.com/fastai/fastai} implementation of ULMFiT. Table \ref{tab:perplexity} shows perplexity of the trained models on validation set. Pre-trained models along with datasets can be downloaded and used for several other downstream classification tasks from the GitHub repository.

\begin{table}[]
\centering
\caption{Validation set perplexity of ULMFiT, trained on synthetically generated code-mixed data}
\label{tab:perplexity}
\begin{tabular}{cccc}
\hline
\textbf{Language} & \textbf{Tamil} & \multicolumn{2}{c}{\textbf{Malayalam}} \\ \hline
\textbf{Script} & Latin & Latin & Mixed \\
\textbf{Perplexity} & 37.50 & 45.84 & 41.22 \\ \hline
\end{tabular}
\end{table}

\textbf{Preprocessing of Dravidian code-mix HASOC dataset.} Basic preprocessing was done by lower-casing, removing @username mentions etc. Exact reproduction details are available in the GitHub repository.

\textbf{Classifier training.} Classifier was trained by first fine-tuning the pre-trained ULMFiT language model and then training the classifier.

\section{Experiment and Results}

\subsection{Experiment setting}
We did basic pre-processing detailed in the previous section. For each one of the sub-tasks, we split the training data into standard train-validation splits in the ratio of 80:20, in case explicit validation set was not given. Using this split, we perform experiments to identify the best set of hyperparameters, shown in Table \ref{tab:hyperparams-classifier}. Gradual unfreezing epochs and learning rates are accessible from notebooks in GitHub repository. Backpropagation through time (BPTT) was set to 70. For language model fine-tuning, dropout multiplicity was set to 0.3, batch size was set to 64, the model was trained for 1 epoch with learning rate 1e-2 with only last layer unfreezed and 5 epochs with learning rate 1e-3 after unfreezing all layers.

\begin{table}[]
\centering
\caption{Hyperparameters for Classifier training}
\label{tab:hyperparams-classifier}
\begin{tabular}{cccccc}
\hline
\multicolumn{2}{c}{\textbf{Task}} & \textbf{\begin{tabular}[c]{@{}c@{}}Dropout\\ Multiplicity\end{tabular}} & \textbf{\begin{tabular}[c]{@{}c@{}}Batch\\ Size\end{tabular}} & \multicolumn{2}{c}{\textbf{\begin{tabular}[c]{@{}c@{}}Model\\ Unfreezed\end{tabular}}} \\ \cline{5-6} 
\textbf{} &  & \textbf{} &  & \textbf{Epochs} & \textbf{Learning Rate} \\ \hline
Sub-Task A & \begin{tabular}[c]{@{}c@{}}Malayalam\\ Mixed Script\end{tabular} & 0.5 & 16 & 5 & 1e-3 \\ \hline
\multirow{2}{*}{Sub-Task B} & Malayalam & 0.7 & 16 & 5 & 1e-3 \\
 & Tamil & 0.5 & 16 & 5 & 1e-3 \\ \hline
\end{tabular}
\end{table}

\subsection{Results}
In this competition, teams were ranked by the weighted F1 score of their classification system. Tables \ref{tab:task1}, \ref{tab:task2-mal} and \ref{tab:task2-tam} show Top-5 ranked teams in the competition with their weighted Precision, weighted Recall and weighted F1-score for each one of the sub-tasks. Difference between top ranked team's weighted F1-score and our weighted F1-score is less than 0.04 in all the sub-tasks. Perplexity of ULMFiT language model after fine-tuning was 57.53 for mixed-script Malayalam in sub-task A, 71.02 for code-mixed Tamil and 89.12 for code-mixed Malayalam in Latin script in sub-task B.


\useunder{\uline}{\ul}{}
\begin{table*}[]
\centering
\caption{Top-5 of the official leader-board in Sub-task A for code-mixed Malayalam}
\label{tab:task1}
\begin{tabular}{lllll}
\hline
\textbf{TeamName} & \textbf{Precision} & \textbf{Recall} & \textbf{F-Score} & \textbf{Rank} \\ \hline
SivaSai@BITS, IIITG-ADBU & 0.95 & 0.95 & 0.95 & 1 \\
CFILT\_IITBOMBAY, SSNCSE-NLP & 0.94 & 0.94 & 0.94 & 2 \\
CENMates, NIT-AI-NLP, YUN, Zyy1510 & 0.93 & 0.93 & 0.93 & 3 \\
{\textbf{Gauravarora}} & {\textbf{0.92} } & {\textbf{0.91}} & {\textbf{0.91}} & {\textbf{4}} \\
WLV-RIT & 0.89 & 0.9 & 0.89 & 5 \\ \hline
\end{tabular}
\end{table*}

\begin{table*}[]
\centering
\caption{Top-5 of the official leader-board in Sub-task B for code-mixed Malayalam}
\label{tab:task2-mal}
\begin{tabular}{lllll}
\hline
\textbf{TeamName} & \textbf{Precision} & \textbf{Recall} & \textbf{F-Score} & \textbf{Rank} \\ \hline
CENmates & 0.78 & 0.78 & 0.78 & 1 \\
SivaSai, KBCNMUJAL & 0.79 & 0.75 & 0.77 & 2 \\
IIITG-ABDU & 0.77 & 0.76 & 0.76 & 3 \\
SSNCSE-NLP & 0.78 & 0.74 & 0.75 & 4 \\
{\textbf{Gauravarora}} & {\textbf{0.76}} & {\textbf{0.72}} & {\textbf{0.74}} & {\textbf{5}} \\ \hline
\end{tabular}
\end{table*}

\begin{table}[]
\centering
\caption{Top-5 of the official leader-board in Sub-task B for code-mixed Tamil}
\label{tab:task2-tam}
\begin{tabular}{lllll}
\hline
\textbf{TeamName} & \textbf{Precision} & \textbf{Recall} & \textbf{F-Score} & \textbf{Rank} \\ \hline
SivaSaiBITS & 0.9 & 0.9 & 0.9 & 1 \\
{\textbf{Gauravarora}} & {\textbf{0.88}} & {\textbf{0.88}} & {\textbf{0.88}} & {\textbf{2}} \\
SSNCSE-NLP & 0.88 & 0.88 & 0.88 & 2 \\
KBCNMUJAL, IIITG-ADBU, zyy1510 & 0.87 & 0.87 & 0.87 & 3 \\
CENmates, CFILT & 0.86 & 0.86 & 0.86 & 4 \\
YUN & 0.85 & 0.85 & 0.85 & 5 \\ \hline
\end{tabular}
\end{table}

\section{Conclusion}

In this paper, we presented a Markov model to facilitate synthetic generation of code-mixed data and used it to pre-train ULMFiT language model on Dravidian languages. Using transfer learning with the pre-trained ULMFiT model on downstream task of hate speech detection, we achieved Rank 2 for Tamil in Sub-task B, Rank 4 for Malayalam in Sub-task A and Rank 5 for Malayalam in Sub-task B. In future research, we will consider improving synthetically generating code-mixed data by allowing code-mixing within a single sentence and using Transformer based models.

\bibliography{sample-ceur}

\end{document}